\documentclass[letterpaper, 10 pt, journal, twoside]{IEEEtran}

\IEEEoverridecommandlockouts                              

\usepackage{geometry}
\usepackage{graphicx}
\usepackage{hyperref}
\usepackage{multirow}
\usepackage{amsmath}
\usepackage{amsmath,amssymb}
\usepackage{bbm}
\usepackage{tcolorbox}
\usepackage{colortbl}
\usepackage{xcolor}

\usepackage{caption}
\usepackage{subcaption}
\usepackage{booktabs}
\usepackage{geometry}

\usepackage{placeins}

\definecolor{light-gray}{HTML}{BFBFBF}
\definecolor{soft-yellow}{HTML}{F7E463}
\definecolor{vibrant-orange}{HTML}{E69F00}

\newcommand{\graybg}[1]{%
  \begingroup\setlength{\fboxsep}{0pt}
  \colorbox{light-gray}{#1}%
  \endgroup
}

\newcommand{\yellowbg}[1]{%
  \begingroup\setlength{\fboxsep}{0pt}
  \colorbox{soft-yellow}{#1}%
  \endgroup
}

\newcommand{\orangebg}[1]{%
  \begingroup\setlength{\fboxsep}{0pt}
  \colorbox{vibrant-orange}{#1}%
  \endgroup
}

\cellcolor[HTML]{F7E463} 
\cellcolor[HTML]{BFBFBF} 
\cellcolor[HTML]{E69F00} 

\usepackage{tablefootnote}
\usepackage{threeparttable}

\usepackage[numbers, sort&compress]{natbib} 

\usepackage{titlesec}
\titlespacing{\section}{0pt}{2pt}{2pt} 
\titlespacing{\subsection}{0pt}{1pt}{0pt}
\titlespacing{\subsubsection}{0pt}{0pt}{0pt}

\usepackage{algorithm}
\usepackage{algpseudocode}
\usepackage{needspace}

\usepackage{svg}
\usepackage{dsfont}

\usepackage{amssymb}  

\begin{document}
\title{Strip-Fusion: Spatiotemporal Fusion for Multispectral Pedestrian Detection}

\author{Asiegbu Miracle Kanu-Asiegbu$^{1}$, Nitin Jotwani$^{2}$, and Xiaoxiao Du$^{3}$
\thanks{Manuscript received: July, 16, 2025; Revised November, 23, 2025; Accepted January, 5, 2026. 
This paper was recommended for publication by Editor Pascal Vasseur upon evaluation of the Associate Editor and Reviewers' comments.
This work was supported by the National Science Foundation under Grant IIS-2153171-CRII: III: Explainable Multi-Source Data Integration with Uncertainty. A. M. Kanu-Asiegbu is supported by a Rackham Merit Fellowship.}

\thanks{$^{1}$A. M. Kanu-Asiegbu is with the Department of Mechanical Engineering, University of Michigan, Ann Arbor, MI 48109 USA  {\tt\small akanu@umich.edu}}%
\thanks{$^{2}$N. Jotwani is with the Electrical Engineering and Computer Science Department, University of Michigan, Ann Arbor, MI 48109 USA {\tt\small njotwani@umich.edu}}
\thanks{$^{3}$X. Du is with the Robotics Department, University of Michigan, Ann Arbor, MI 48109 USA {\tt\small xiaodu@umich.edu}}%

\thanks{Digital Object Identifier (DOI): see top of this page. Code and supplementary multimedia materials available at \url{https:/github.com/akanuasiegbu/stripfusion}.}
}

\markboth{IEEE Robotics and Automation Letters. Preprint Version. Accepted January, 2026}
{Kanu-Asiegbu \MakeLowercase{\textit{et al.}}: Strip-Fusion}

\maketitle

\begin{abstract}
Pedestrian detection is a critical task in robot perception. Multispectral modalities (visible light and thermal) can boost pedestrian detection performance by providing complementary visual information. Several gaps remain with multispectral pedestrian detection methods. First, existing approaches primarily focus on spatial fusion and often neglect temporal information. Second, RGB and thermal image pairs in multispectral benchmarks may not always be perfectly aligned. Pedestrians are also challenging to detect due to varying lighting conditions, occlusion, etc. This work proposes \textit{Strip-Fusion}, a spatial-temporal fusion network that is robust to misalignment in input images, as well as varying lighting conditions and heavy occlusions. The \textit{Strip-Fusion} pipeline integrates temporally adaptive convolutions to dynamically weigh spatial-temporal features, enabling our model to better capture pedestrian motion and context over time. A novel Kullback–Leibler divergence loss was designed to mitigate modality imbalance between visible and thermal inputs, guiding feature alignment toward the more informative modality during training. Furthermore, a novel post-processing algorithm was developed to reduce false positives. Extensive experimental results show that our method performs competitively for both the KAIST and the CVC-14 benchmarks. We also observed significant improvements compared to previous state-of-the-art on challenging conditions such as heavy occlusion and misalignment.

\end{abstract}


\vspace{-1mm}
\section{Introduction}
\vspace{-1mm}

\IEEEPARstart{P}{edestrian} detection is an important task in computer vision and robot perception. In applications such as autonomous driving, accurate detection of pedestrians helps ensure safe and efficient path planning and human-vehicle interactions. Multispectral pedestrian detection using visible light (RGB) and thermal modalities has gained interest in recent years \cite{ha2017mfnet, kim2021uncertainty, jiang2025research}. While using the RGB modality alone can be vulnerable to challenging environments, such as low light conditions, the thermal modality provides complementary information and has been shown to boost pedestrian detection performance \cite{kim2021mlpd, liu2024fdenet}. 

Several challenges exist for multispectral pedestrian detection. First, existing approaches typically only consider fusing static RGB and thermal image pairs and often ignore the temporal information \cite{qingyun2021cross, li2025multimodal}. However, pedestrians move in periodic motion \cite{ran2007pedestrian}, and it would be beneficial to take temporal features into consideration.  Second, pedestrians in urban environments are highly dynamic and cluttered \cite{wu2016robust}. Pedestrians are also challenging to detect in poorly lit or obscured environments. Third, in multispectral object detection, there often exists disproportionate contribution of one modality on the detection results (called ``modality imbalance'') \cite{zhou2020improving}. Intuitively, the RGB images have clearer texture features during daytime, whereas the thermal modality can detect pedestrian shapes more easily in nighttime/low-light conditions. It is necessary to incorporate sufficient cross-modality features to adapt to illumination and feature variations. Fourth, spatial misalignment is common in multispectral pedestrian datasets due to different field-of-view between RGB and thermal cameras, as well as miscalibration \cite{kim2021uncertainty}. Thus, instead of directly fusing the same region of interest for a pedestrian, a robust approach is needed to handle uncertainty in pedestrian spatial locations.

This paper proposes a novel spatial-temporal multispectral fusion approach, \textit{Strip-Fusion}, to address the above challenges. To help the model implicitly learn spatial-temporal associations, Temporally Adaptive Convolutions (TAdaConv) \cite{huang2021tada} were used to enhance feature representations.  Our fusion network extends the mixing modules of Strip-MLP \cite{cao2023strip}, a model inspired from the MLP-Mixer family \cite{tolstikhin2021mlp}, to work with multispectral inputs while effectively capturing both long-range and local spatial dependencies.
In addition, a novel Kullback–Leibler (KL) divergence loss was proposed to reduce modality imbalance. We show later in ablation studies that our KL divergence loss is particularly effective for short temporal windows by encouraging alignment between visible and thermal feature distributions. Furthermore, \textcolor{black}{to help improve robustness of our detections,} a novel post-processing algorithm was designed to address inconsistencies between visible and thermal detection heads and we show that it effectively improved detection performance. Experimental results were presented on two challenging multispectral pedestrian detection benchmarks, KAIST \cite{hwang2015multispectral} and CVC-14 \cite{gonzalez2016pedestrian}. Our proposed method achieves competitive results compared to the state-of-the-art methods, and we observed significant performance improvement on challenging cases such as heavy occlusions (KAIST) and spatial misalignment (CVC-14).

Our contributions include:
%
\begin{itemize}
\item A novel spatial-temporal fusion method,  \textit{Strip-Fusion}, for multispectral pedestrian detection. We incorporate temporally adaptive convolutions and an MLP-based temporal fusion module to learn spatial-temporal features, and extend Strip Mixing Modules for cross-spectral fusion.
\item A novel KL-Divergence loss to address modality imbalance by encouraging feature distribution alignment between RGB and thermal modalities.
\item A novel post-processing strategy at the feature-map scale to match pedestrian detection pairs between modalities and reduce false positives.
\item Experimental results on the KAIST and CVC-14 benchmarks demonstrate competitive performance of our proposed  \textit{Strip-Fusion} compared to the state-of-the-art. We achieve superior performance on challenging scenarios such as heavy occlusion and spatial misalignment.
\item We also provide extensive ablation studies and analysis on the effects of various components and hyperparameters.
\end{itemize}

\section{Related Work}
\subsection{Multispectral Pedestrian Detection}

Classic fusion approaches for multispectral pedestrian detection used handcrafted features \cite{hwang2015multispectral, gonzalez2016pedestrian}. Since the rise of deep learning, researchers have developed fusion approaches using architectures such as convolutional neural networks \cite{kim2021mlpd, zhou2020improving, zhang2019weakly, kim2021uncertainty, kim2024causal} and Transformers \cite{xing2024ms, lee2022cross, qingyun2021cross}. Transformer-based fusion approaches typically have quadratic complexity with respect to the input size, and require downsampling of feature maps to reduce computational cost \cite{lee2022cross,qingyun2021cross}. Mamba \cite{gu2024mamba}, a state-space-based model with linear complexity with respect to input size, has also been adapted and used for multispectral fusion \cite{gao2024mambast, dong2025fusion, li2024cfmw}. Our method \textit{Strip-Fusion} is based on multi-layer perceptrons (MLPs), which have a simple architecture and linear complexity with respect to input size.

\subsection{Modality Imbalance}
In multispectral object detection, modality imbalance refers to the disproportionate contribution of one modality, often thermal, on the detection results, which can limit the effectiveness of fusion. 
Kim et al. \cite{kim2024causal} demonstrated that existing multispectral detection methods struggle when pedestrians are identifiable in the visible spectrum but hidden in the thermal spectrum, such as when a pedestrian is wearing a heat-insulating jacket, highlighting the bias towards the thermal spectrum. Some approaches that attempt to deal with modality imbalance are illumination-aware fusion \cite{zhou2020improving}, confidence-aware fusion \cite{ zhang2019weakly}, KL divergence \cite{kim2021uncertainty}, and causal inference  \cite{kim2024causal}. In this work, we adopted the KL divergence for its simplicity and implicit incorporation of illumination variation through model uncertainty. 

\subsection{Handling Spatial Misalignment}
Spatial misalignment in visible and thermal image fusion is common due to the cameras not being co-located, or hardware discrepancies such as differences in camera resolution and fields-of-view \cite{hwang2015multispectral, gonzalez2016pedestrian}. 
Most existing multispectral pedestrian detection methods focus on dealing with ``weakly aligned'' images, where RGB and thermal image pairs that are not perfectly aligned pixel-wise, but regarded as ``close enough'' to enable reasonable detection \cite{zhang2019weakly}.
AR-CNN \cite{zhang2019weakly} addressed weak misalignment by randomly shifting the object proposals in the visible modality and  aligning the region features between modalities.
MBNET \cite{zhou2020improving} obtained object proposals and used deformable convolutional neural networks (CNNs) to help refine and align the proposals. Kim et al. \cite{kim2021uncertainty} evaluated ambiguities in object regions of interest (RoI) and used predictive uncertainty to measure how reliable the detector is for a given RoI feature. An uncertainty-aware guiding module was proposed to alleviate  uncertainties and address modality discrepancy.
To address heavier misalignment, Wanchaitanawong et al. \cite{wanchaitanawong2021multi} used a region proposal network for each modality to obtain object proposal pairs, which are then refined separately by two detection heads, one visible and one thermal, with each proposal assigned to only one detection head. Inspired by \cite{wanchaitanawong2021multi}, our model uses two detection heads to learn modality-specific detections, and then applies a novel post-processing strategy, which helps further refine and match pedestrian detections and reduce the impact of spatial misalignment.

\begin{figure}[t!]
\includegraphics[width=\columnwidth]{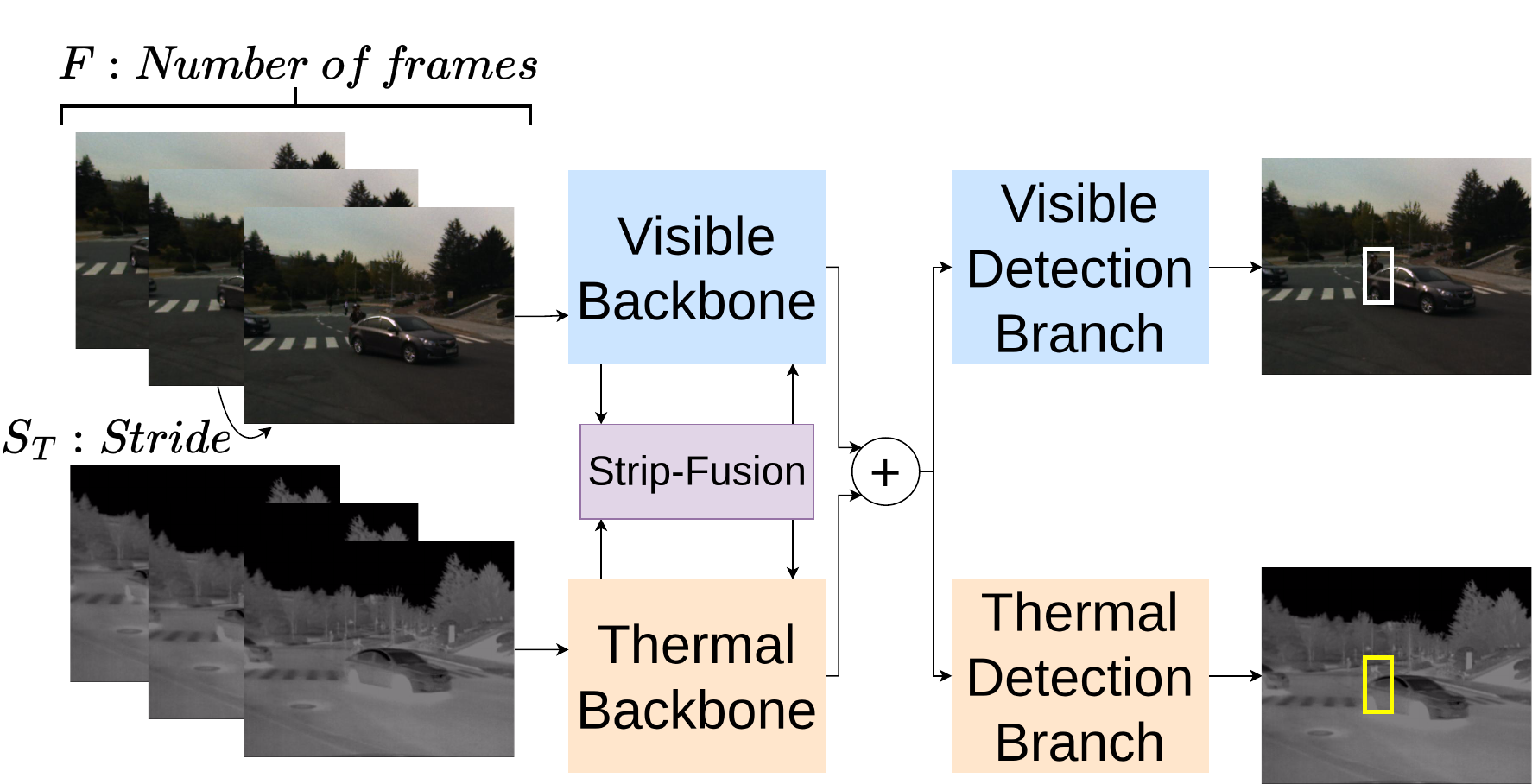}
\vspace{-6mm}
\caption{
\textcolor{black}{High-level overview of \textit{Strip-Fusion}. Input: Sequence of visible–thermal image pairs, each image treated as a 3-channel input. Output: Detected pedestrian bounding boxes at the final frame. Strip-Fusion combines visible and thermal features at multiple scales: 80$\times$80, 40$\times$40, and 20$\times$20.}
}
\label{fig:method_overview}
\vspace{-6mm}
\end{figure}

\section{Method}

\subsection{Overview}
\textcolor{black}{
Our method is a two-stage training strategy. In Stage~1, we learn spatial unimodal models for visible and thermal images independently. These unimodal features are used to initialize our spatial-temporal multimodal model. In Stage~2 , the focus of this paper, we introduce \textit{Strip-Fusion}, a novel spatial-temporal fusion network that combines visible and thermal information across multiple frames to detect pedestrian bounding boxes on the last (current) frame. As shown in Fig.~\ref{fig:method_overview}, our detection pipeline consists of modality-specific backbones, modality-specific detection branches, and the strip fusion module. Given a visible-thermal sequence, frame-wise features are extracted independently and then fused at multiple scales by the novel strip fusion module. The fused features are added back into each detection branch. During Stage 2, we add temporally adaptive convolutions (TAdaConv) \cite{huang2021tada} to the modality-specific backbones and detection branches. TAdaConv extends the spatial unimodal features (``base weights") with temporal modeling, allowing the network to build spatial-temporal representations. Combining TAdaConv with our strip fusion module, enables our model to effectively learn both spatial-temporal and multimodal information.}

\begin{figure*}[t!]
\includegraphics[width=\textwidth]{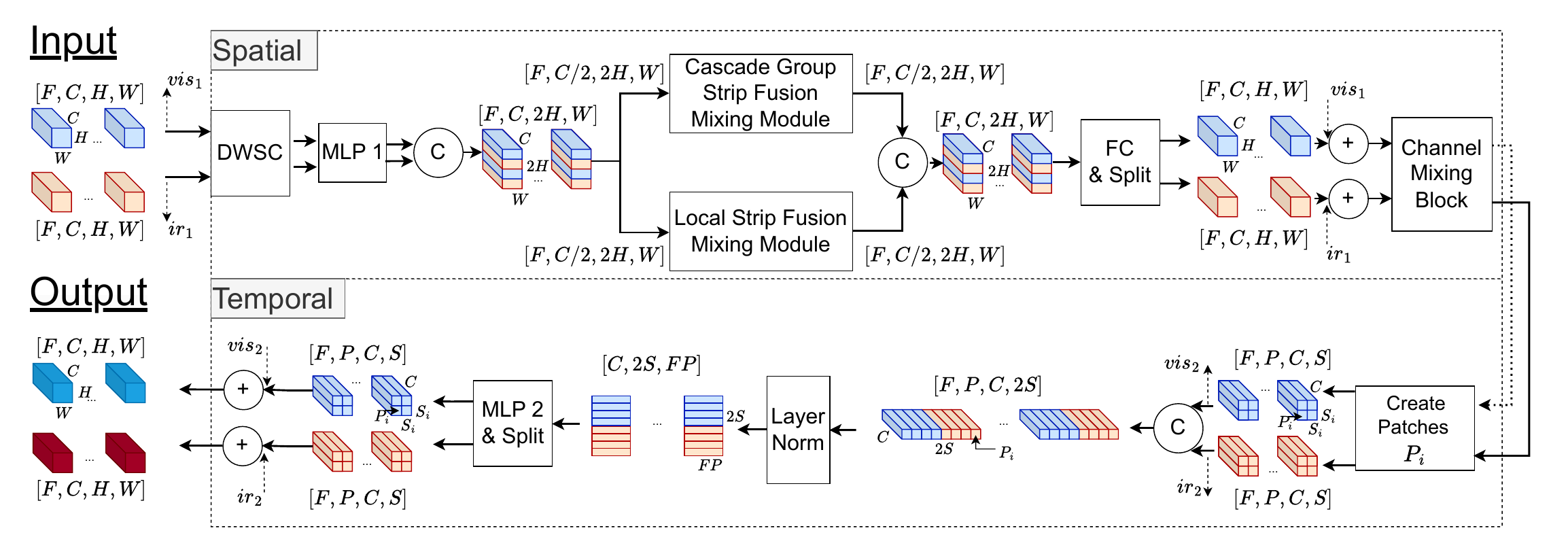}
\vspace{-7mm}
\caption{Our proposed strip fusion module for spatial and temporal fusion. Best viewed in color.}
\label{fig:fusion_overview}
\vspace{-7mm}
\end{figure*}

\begin{figure*}[t!]
\centering
\includegraphics[width=\textwidth]{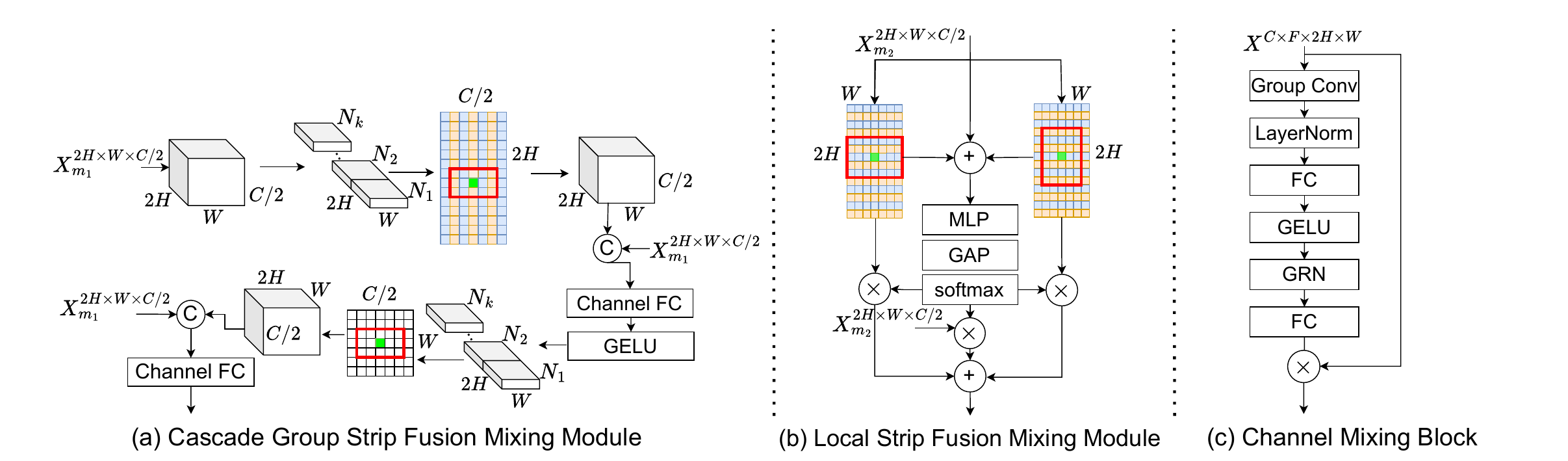}
\vspace{-7mm}
\caption{
(a) Given input feature $X_{m_1}$ $\in$   $\mathbb{R}^{ 2H \times W \times C/2}$, the CGSFMM splits the feature into $N$ patches along the channel direction, and applies Strip-MLP along the row and column direction in series. (b) Given input feature $X_{m_2}$ $\in$   $\mathbb{R}^{ 2H \times W \times C/2}$, the LSFMM applies a row and column Strip-MLP, combines their outputs, processes the result through an MLP, performs global average pooling, and finally uses a softmax layer to weight the contributions of the row and column components along with the original input. (c) The channel mixing block integrates information across channels and patches. GRN refers to Global Response Normalization. }
\label{fig:fusion_and_channel_mixing}
\vspace{-6mm}
\end{figure*}

\subsection{Strip Fusion Module}
\label{sec:stripfusion}

A new strip fusion module is proposed to integrate multispectral features across spatial and temporal dimensions. 


\subsubsection{Spatial Fusion} Fig.~\ref{fig:fusion_overview} shows a diagram of our proposed strip fusion module. The inputs are sequences of visible and thermal features from backbone, each of size $\mathbb{R}^{F \times C \times H \times W}$, where $F$ is the number of frames, $C$ is the number of channels, $H$ and $W$ are the feature height and width. The modality-specific features are first processed independently using a Depth-Wise Separable Convolution (DWSC) \cite{chollet2017xception} and then a multi-layer perceptron (MLP). Next, the features are  concatenated along the feature height dimension, where the concatenated rows alternate between the visible (blue) and thermal (orange) features .

The concatenated features of size $\mathbb{R}^{F \times C \times 2H \times W}$ are then split into two halves along  the channel dimension. The first half is passed to a Cascade Group Strip Fusion Mixing Module (CGSFMM), while the second half is passed to a Local Strip Fusion Mixing Module (LSFMM). Similar to \cite{cao2023strip}, the CGSFMM aims to model long-range interactions, while the  LSFMM is designed to model local interactions. Outputs from CGSFMM and LSFMM are concatenated along the channel dimension and passed into a fully connected layer, after which the features are split and added with a residual connection of visible \textcolor{black}{($vis_1$)} and thermal 
\textcolor{black}{($ir_1$)} features. Finally, a channel mixing block ensures features interact along $C$. 

Figs. \ref{fig:fusion_and_channel_mixing}a, \ref{fig:fusion_and_channel_mixing}b, and \ref{fig:fusion_and_channel_mixing}c show the CGSFMM,  LSFMM, and channel mixing block. Compared to the original Strip-MLP \cite{cao2023strip}, we added alternating concatenation and increased the convolution kernel sizes (see Sec.~\ref{sec:implementation_details}) to enable fusion, as larger kernel sizes capture more spatial contextual information.

\subsubsection{Temporal Fusion}
\textcolor{black}{ Fig.~\ref{fig:fusion_overview} shows MLP 2, which is responsible for performing temporal fusion. To combine features temporally, the height and width dimensions are first divided into patches of size $s \times s$, producing $P = \frac{HW}{s^{2}}$ patches. After patching, the visible and thermal features each have size $\mathbb{R}^{F \times P \times C \times S}$, where $S$ denotes the feature dimension of each patch. The two modalities are then concatenated along $S$, resulting in features of size $\mathbb{R}^{F \times P \times C \times 2S}$, which is subsequently normalized by a layer norm operation. To allow MLP~2 to fuse information across time, the features are rearranged by merging the frame and patch dimensions, yielding $\mathbb{R}^{C \times 2S \times FP}$. After temporal fusion by MLP~2, the fused representation is split back into visible and thermal components, rearranged to their original structure, and added to their respective residual features ($vis_2$ and $ir_2$).}


\subsection{KL-Divergence}


\textcolor{black}{Inspired by  \cite{kim2021uncertainty}, we introduce a Kullback–Leibler (KL) divergence loss to mitigate modality imbalance by transferring information from the more reliable modality to the less reliable one by encouraging predictions to align. We have two key differences with \cite{kim2021uncertainty}. First, instead of estimating modality uncertainty using Monto Carlo Dropout, we measure modality reliability directly using Complete Intersection over Union (CIoU) \cite{zheng2021enhancing}. Second, to incorporate spatial-temporal context, we extract  region-of-interest (RoI) features across the entire sequence of image pairs. Since ground-truth boxes are available only in the final frame, we select the modality with the higher CIoU in the final frame as the reference, and apply KL divergence to transfer information from the stronger modality to the weaker one, improving cross modality alignment. Below is our KL divergence formulation for one iteration at a single scale. Our model performs multi-scale detection.}

To filter detections, we use the CIoU metric to select the top $N$ bounding boxes closest to the ground truth, where $N = 300$ in our implementation to ensure consistent evaluation. Using these filtered detections, we assess which modality, visible or thermal, produces bounding boxes more closely aligned with the ground truth. This modality is considered more reliable. The reliabilities of the visible ($r^v$) and thermal ($r^t$) modalities are an average of top-$n$ CIoU scores, where $\textit{CIoU}^{v}_n$ and $\textit{CIoU}^{t}_n$ represent the $n$-th top score for each modality, respectively:

\vspace{-5mm}
\begin{equation}
r^{v} = \textstyle \frac{1}{N} \sum_{n=0}^{N-1} \textit{CIoU}^{v}_n, \quad
r^{t} = \textstyle  \frac{1}{N} \sum_{n=0}^{N-1} \textit{CIoU}^{t}_n.
\end{equation}

Given the reliability ($r^v$ and $r^t$), we determine the visible ($f^{v}_n$) and thermal ($f^{t}_n$) features using \textit{RoIAlign} \cite{he2017mask} as
%
\begin{equation}
f^{v}_n = \textit{ROIAlign}(B^{r}_{n}, F^{v}), \quad
f^{t}_n = \textit{ROIAlign}(B^{r}_{n}, F^{t})
\end{equation}
%
where $B^r_n$ represents the $n$-th top bounding box from the reliable modality, $F^v$ and $F^t$ represent the feature map of visible and thermal modality.  The extracted features ($f_n^v, f_n^t \in  \mathbb{R}^{F \times C \times 3 \times 3}$) are flattened into one-dimensional vectors. To compute the visible ($C^{v} \in R^{N \times N}$) and thermal ($C^{t} \in R^{N \times N}$) cosine similarity matrices, we define the visible ($c^{v}_{ij}$) and thermal ($c^{t}_{ij}$) cosine similarity matrix elements as 
%
\vspace{-2mm}
\begin{equation}
c^{v}_{ij} =  \frac{f^v_i \cdot f^v_j}{\|f^v_i\|_2 \cdot \|f^v_j\|_2},\quad
c^{t}_{ij} =  \frac{f^t_i \cdot f^t_j}{\|f^t_i\|_2 \cdot \|f^t_j\|_2}, \quad
\end{equation}
%
where $\cdot$ is the dot product and $\| . \|_2$ is the L2-norm. The coefficients $i$ and $j$ are the row and column indices of the matrix, respectively.

Similar to \cite{kim2021uncertainty}, we compute feature relation matrices $M^v$ and $M^t$ to capture modality-wise feature distributions. The matrix elements are obtained using a row-wise softmax as
\vspace{-2mm}
\begin{equation}
    m_{ij}^{v} = \frac{e^{c_{ij}^{v}}}{\sum_{k=0}^{N-1} e^{c_{ik}^{v}}}, \quad
    m_{ij}^{t} = \frac{e^{c_{ij}^{t}}}{\sum_{k=0}^{N-1} e^{c_{ik}^{t}}}.
\end{equation}

Next, we use KL divergence to push the feature distributions with lower reliability towards the feature distributions with higher reliability, which encourages spatial alignment of the two modalities. Row-wise KL divergence is computed as 
\vspace{-2mm}

\begin{gather}
D_{KL}(M_i^{t} \| M_i^{v}) = \textstyle \sum_{j=0}^{N-1} m_{ij}^{t} \log \frac{m_{ij}^{t}}{m_{ij}^{v}}, \\
D_{KL}(M_i^{v} \| M_i^{t}) = \textstyle \sum_{j=0}^{N-1} m_{ij}^{v} \log \frac{m_{ij}^{v}}{m_{ij}^{t}}.
\end{gather}
%
%
\noindent where $D_{KL}(M_i^{t} \| M_i^{v})$ encourages the visible distribution to align with the thermal, and similarly, $D_{KL}(M_i^{v} \| M_i^{t})$ encourages the thermal distribution to align with the visible.

The KL divergence loss function ($\mathcal{L}_{\text{KL}}$) is written as
\begin{equation}
    \begin{split}
        \mathcal{L}_{\text{KL}}  = &
        \; \mathds{1}[r^{t} > r^{v}] 
        \textstyle \sum_{i=0}^{N-1} D_{KL}(M_i^{t} \| M_i^{v}) \\
        & + 
         (1-  \mathds{1}[r^{t} > r^{v}]) 
        \textstyle \sum_{i=0}^{N-1} D_{KL}(M_i^{v} \| M_i^{t}),
    \end{split}
\end{equation}

\noindent where  $\mathds{1}[.]$ is the indicator function determining which modality is more reliable.

\subsection{Total Loss}

The final total loss function is written as 
\begin{equation}
     \mathcal{L}_{tot} = \mathcal{L}_{reg}^{v} + \mathcal{L}_{reg}^{t} + \mathcal{L}_{obj}^{v}  + \mathcal{L}_{obj}^{t}  + \beta \mathcal{L}_{\text{KL}},
\end{equation}
\noindent where $\mathcal{L}_{reg}^{v}$ and $\mathcal{L}_{reg}^{t}$ are the visible and thermal regression losses, $\mathcal{L}_{obj}^{v}$ and $\mathcal{L}_{obj}^{t}$ are the visible and thermal objectness losses, and $\beta$ is a hyperparameter controlling the influence of the Kullback–Leibler (KL) divergence loss. 

\vspace{-2mm}
\begin{algorithm}[H]
\caption{Post-processing at Each Feature-Map Scale}
\label{alg:fuse_predictions}
\begin{algorithmic}[1]
\Require $B^v, B^t, s^v, s^t, iou_{\text{thres}}$

\State $\text{fused\_bbox} \gets [\;]$   \Comment{Initialize fused detections}

\For{each batch in $(B^v, B^t)$}

    \State $B^k_{\text{fil}} \gets \text{Filter}(B^k, s^k), \quad k \in \{v, t\}$

    \State \textbf{if} $B_{\text{fil}}^v = \emptyset$ \textbf{or} $B_{\text{fil}}^t = \emptyset$ \textbf{then continue}

    \State $match\_pairs \gets 
    \left\{(b^v, b^t)\ \middle|\ 
    \text{IoU}(B^v_{\text{fil}}, B^t_{\text{fil}}) \ge iou_{\text{thres}}\right\}$

    \For{each $(b^v, b^t) \in match\_pairs$}
        \State $(x_{\min}, y_{\min}) \gets \min(b^v_{tl}, b^t_{tl})$
        
        \State $(x_{\max}, y_{\max}) \gets \max(b^v_{br}, b^t_{br})$

        \State $(x_c, y_c) \gets (x_{\min}+x_{\max},\, y_{\min}+y_{\max})/2$
        
        \State $(w, h) \gets (x_{\max}-x_{\min},\, y_{\max}-y_{\min})$
        
        \State $f_{\text{conf}} \gets (s^v + s^t) / 2$
        
        \State $\text{fused\_bbox} \gets \text{fused\_bbox} \,\Vert\, (x_c, y_c, w, h, f_{\text{conf}})$

    \EndFor
\EndFor

\State \textbf{return} $\text{fused\_bbox}$
\end{algorithmic}
\end{algorithm}
\vspace{-4mm}
\subsection{Post Processing}
\textcolor{black}{We introduce a new postprocessing approach, Algorithm~\ref{alg:fuse_predictions}, a late stage fusion strategy designed to suppress detections arising from weak or inconsistent cross-modal signals.}
This postprocessing algorithm is carried out independently at each feature scale, ensuring that only detections associated with anchors of corresponding sizes are grouped and processed. At a single feature scale, given the bounding box detections from the visible ($B^v$) and thermal ($B^t$) modalities, we first filter detections based on their respective confidence scores ($s^v$ for visible and $s^t$ for thermal). Next, we compute the Intersection over Union (IoU) between all filtered pairs of visible and thermal bounding boxes. Then, we identify matched pairs $(b^v, b^t)$, based on a predefined IoU threshold. Once pairs are determined, we update the bounding box locations using the convex hull of each pair. The updated confidence score for a single pair, $f_{\text{conf}}$, is calculated as the average of the visible and thermal confidence scores, rather than the maximum, as this yielded better performance in our experiments \textcolor{black}{(See Multimedia
Tables R1 \& R2)}. The post-processed bounding boxes and confidence scores are appended (using the operator $\Vert$) and returned for the current feature scale, and the same procedure is repeated for each additional feature scale.

\section{Experiments}
\subsection{Datasets and Evaluation Metric}
Experiments are conducted on two benchmark multispectral pedestrian detection datasets, KAIST \cite{hwang2015multispectral} and CVC-14 \cite{gonzalez2016pedestrian}. Both datasets contain image sequences collected from visible and thermal cameras mounted on a moving vehicle, providing ego-centric videos of pedestrians in urban driving settings.

KAIST \cite{hwang2015multispectral} is a multispectral dataset containing weakly thermal and RGB images. The original dataset contains about 95,000 images, collected at 20 Hz, with a resolution of 640$\times$512 \textcolor{black}{pixels}. However, the raw annotations were found to be noisy, and Li et al. \cite{li_2018_BMVC} reduced annotation noise by providing so-called ``sanitized annotations" for training on 7601 image pairs, which we use in this work. For fair comparisons with state-of-the-art methods, a standard testing set containing 2252 images is used, of which 1455 are daytime images and 797 are nighttime images. Three occlusion tags of pedestrians are given 1) none, 2) partial, and 3) heavy.


CVC-14 \cite{gonzalez2016pedestrian} is a multispectral dataset captured at a rate of 10 Hz, with images at a resolution of 640$\times$471 pixels. It contains misaligned and weakly aligned thermal and grayscale image pairs. 
Due to misalignment, separate bounding box labels are provided for each modality. The dataset is divided into 7085 training and 1417 testing image pairs (690 daytime and 727 nighttime images). For training, we use only RGB labels, each of which has a corresponding thermal pair. During inference,  to ensure fair comparison with state-of-the-art methods, only RGB labels are used. No occlusion tags given.


We use log-average miss rate (MR) as our primary evaluation metric, following standard practice in pedestrian detection~\cite{dollar2011pedestrian}. 
To assess performance under different lighting conditions, we report MR for daytime (MR-Day), nighttime (MR-Night), and overall (MR-All) conditions. The KAIST and CVC-14 datasets are evaluated under the ``Reasonable'' setting, which includes non-occluded or partially occluded pedestrians taller than 55 pixels. For KAIST and CVC-14, we also report results for unoccluded pedestrians at the ``Near" ($>115$ pixels), ``Medium" (45--115 pixels), and ``Far" ($<45$ pixels) ranges. On KAIST, independent of height, we report results for the occlusion levels: ``None" (unoccluded), ``Partial," and ``Heavy". Finally, we include the ``All'' setting, which considers all pedestrians regardless of occlusion or height.


\subsection{Implementation Details}
\label{sec:implementation_details}
For both datasets, our multimodal models were initialized using weights from two YOLOv5 \cite{jocher2020yolov5} models trained on KAIST, one using only RGB images and one using only thermal images. In our work, the weights from YOLOv5 are referred to as base weights \cite{huang2021tada}. On both datasets, we initialized the base weights for the visible backbone and visible detection branch using YOLOv5 trained on KAIST RGB images, while the base weights for the thermal backbone and thermal detection branch are initialized using YOLOv5 trained on KAIST thermal images. 
\textcolor{black}{Base weights were frozen based on the ablation studies (Sec.~\ref{sec:ablation}), though we always train the calibration weights \cite{huang2021tada} to learn temporal information.} The CGSFMM uses a kernel size of $1 \times 5$ for the height and width Strip-MLP. The LSFMM uses $5 \times 7$ for the height Strip-MLP and $7 \times 5$ for the width Strip-MLP. The channel mixing block applies group convolution with a kernel size of $11 \times 11$. 

A batch size of six is used for both datasets. The models were trained using two NVIDIA A100 GPUs. Our models were optimized using stochastic gradient descent (SGD) with a momentum of 0.843 and a Cosine Learning Rate Schedule that decays from 1.0 to 0.12. Additionally, we employ a two-epoch warm-up phase. We train KAIST and CVC-14 models for 25 and 40 epochs, respectively. To mitigate distribution shift, CVC-14 images and annotations were resized to $640\times512$ to match the KAIST dataset during training and inference, but we rescale the annotations back to $640\times471$ for final evaluation.

\addtocounter{footnote}{1} 
\footnotetext[\value{footnote}]{\label{fn:retrain}Indicates retrained models. MS-DETR  retrained on CVC-14 only.}

\begin{table}[t!]
\centering
\renewcommand{\arraystretch}{1.2} 
\caption{{Miss Rate Evaluation on KAIST and CVC-14. \yellowbg{Yellow}, \graybg{Grey}, and \orangebg{Orange} color indicate Best, Second Best, and Third Best. Evaluated on Reasonable Setting. Lower MR is better. $F$: Number of Frames. $S_T$: Stride.}}
\vspace{-0.1cm} 
\resizebox{\columnwidth}{!}{
\begin{tabular}{l|ccc|ccc}
\hline
\multirow{3}{*}{Methods} & \multicolumn{3}{c|}{KAIST }    & \multicolumn{3}{c}{CVC-14} 
\\ \cline{2-7} 
                                        & MR-All   & MR-Day & MR-Night & MR-All & MR-Day &  MR-Night  \\ \hline
YOLOv5 (RGB)                            & 22.25 & 18.95 &	30.81       & 38.31 & 41.19 & 34.63  \\ 
YOLOv5 (IR)                             & 21.83	& 24.83 &	15.69          & 80.77 & 91.88 & 62.99       \\                                 
AR-CNN \cite{zhang2019weakly}           & 9.34  & 9.94  &   8.38             & 22.1 & 24.7 & 18.1         \\ 
CMM \cite{kim2024causal}                & 8.54  & 9.60  &   8.54                     & \cellcolor[HTML]{E69F00}{17.13} & 27.81 & \cellcolor[HTML]{F7E463}{7.71}  \\ 
MBNet \cite{zhou2020improving}          & 8.13  & 8.28  & 7.86               & 21.1 & 24.7 & 13.5 \\ 
Kim et al. \cite{kim2021uncertainty}    & 7.89 & 8.18 & 6.96               & 18.7 & 23.87 & \cellcolor[HTML]{E69F00}{11.08}      \\ 
MLPD \cite{kim2021mlpd}                 & 7.58 & \cellcolor[HTML]{E69F00}{7.95} & 6.95               & 21.33 & 24.18  & 17.97    \\ 
\footnotemark[1] MambaST \cite{gao2024mambast}           & \cellcolor[HTML]{E69F00}{7.26} & 8.98 & 4.43   & 18.20  & 21.91 & 13.87    \\ 
Lee et al. \cite{lee2022cross}          & \cellcolor[HTML]{BFBFBF}{7.03} & \cellcolor[HTML]{F7E463}{7.51} & 6.53        & 20.58  & 23.97  & 13.85     \\ 
\footnotemark[1] MS-DETR \cite{xing2024ms}               & \cellcolor[HTML]{F7E463}{6.13}  &  \cellcolor[HTML]{BFBFBF}{7.78} & \cellcolor[HTML]{F7E463}{3.18}  & 17.15 & \cellcolor[HTML]{E69F00}{21.25}  & \cellcolor[HTML]{BFBFBF}{9.25}  \\ \hline
Ours ($F,S_T=1,1$) & 7.44	& 9.11 & \cellcolor[HTML]{E69F00}{4.34} &18.02 &	23.02	&11.96 \\

Ours ($F,S_T=3,3$) & 8.02& 10.29 & 4.44 & \cellcolor[HTML]{BFBFBF}{16.53} &	\cellcolor[HTML]{F7E463}{18.13} &	12.90            \\ 

Ours ($F,S_T=5,3$) &9.70&	12.78	& \cellcolor[HTML]{BFBFBF}{4.11}& \cellcolor[HTML]{F7E463}{16.34}&	\cellcolor[HTML]{BFBFBF}{19.30}	&11.67               \\ \hline

\end{tabular}
}
\label{table:mr_KAIST_cvc_14}
\vspace{-2mm}
\end{table}

\begin{table}[t!]
\centering
\renewcommand{\arraystretch}{1.2} 
\caption{{Additional KAIST detection results (MR-All).}}
\resizebox{\columnwidth}{!}{
\begin{tabular}{l|c|c|c|c|c|c|c}
\hline
Methods & Near & Medium & Far & None & Partial&Heavy&All  \\ \hline
YOLOv5 (RGB) & 0.00 & 35.74 & 74.16 & 40.82 & 52.93 & 63.58 & 44.90 \\
YOLOv5 (IR) & 2.02 & 31.73 & 66.57 & 37.96 & 44.61 & 66.12  &41.67 \\
AR-CNN \cite{zhang2019weakly} & 0.00 & 16.08 & 69.00 & 31.40 & 38.63 & 55.73 & 34.95\\
MBNet \cite{zhou2020improving}  & 0.00 & 16.09 & 55.99 & 27.75 & 59.73 &  59.14& 31.87  \\
MLPD \cite{kim2021mlpd} & 0.00 & \cellcolor[HTML]{BFBFBF}{11.93} & 50.86 & 24.15 & 28.75 & 53.97 & 28.49 \\
\footnotemark[1] MambaST \cite{gao2024mambast}  & 0.00 & 12.34 & 36.91 & 19.89 & \cellcolor[HTML]{BFBFBF}{24.55} & \cellcolor[HTML]{E69F00}{44.68} & \cellcolor[HTML]{E69F00}{23.75} \\
MS-DETR \cite{xing2024ms} &  0.00 & \cellcolor[HTML]{F7E463}{9.70} & \cellcolor[HTML]{F7E463}{32.41} & \cellcolor[HTML]{F7E463}{16.54} & \cellcolor[HTML]{F7E463}{23.61} & 48.71 & \cellcolor[HTML]{F7E463}{20.64} \\ \hline

Ours ($F,S_T=1,1$) & 0.00 & 12.68 & \cellcolor[HTML]{BFBFBF}{34.41}  & \cellcolor[HTML]{E69F00}{19.38}  & \cellcolor[HTML]{E69F00}{26.65} & 47.38 & 23.89  \\

Ours ($F,S_T=3,3$) & 0.00 & \cellcolor[HTML]{E69F00}{12.25} & \cellcolor[HTML]{E69F00}{35.76} & \cellcolor[HTML]{BFBFBF}{19.17} & 27.78 & \cellcolor[HTML]{F7E463}{41.05} & \cellcolor[HTML]{BFBFBF}{23.09} \\

Ours ($F,S_T=5,3$) & 0.00 & 15.71& 39.37& 22.75& 27.78& \cellcolor[HTML]{BFBFBF}{42.63}&26.21  \\ \hline

\hline
\end{tabular}
}
\label{table:additional_kaist}
\vspace{-7mm} 
\end{table}

 \begin{table}[h!]
  \vspace{-1mm}
 \centering
 \caption{Additional CVC-14 detection results (MR-All).}
 \begin{tabular}{c|c|c|c}
 \hline
  Methods  & Near & Medium & Far  \\ \hline
\footnotemark[1] MambaST \cite{gao2024mambast} & 11.60 & 27.15 & \cellcolor[HTML]{BFBFBF}{57.61} \\
 \footnotemark[1] MS-DETR \cite{xing2024ms} & \cellcolor[HTML]{BFBFBF}{11.30} & \cellcolor[HTML]{BFBFBF}{26.50}   & \cellcolor[HTML]{F7E463}{56.06}                \\   \hline
Ours ($F,S_T=5,3$)  & \cellcolor[HTML]{F7E463}{9.72}  & \cellcolor[HTML]{F7E463}{25.24} & 60.63                  \\ \hline
 \end{tabular}
 \label{tab:additional_cvc14}
 \vspace{-3mm}
 \end{table} 

\begin{table*}[t]
\centering
\renewcommand{\arraystretch}{1.2} 
\caption{
MR-All Results of freezing base weights on KAIST and CVC-14.  Frames and Stride set to three.}
\begin{tabular}{c c c c | c c c c | c c c c}
\hline
\multicolumn{4}{c|}{Freezing Base Weights} & \multicolumn{4}{c|}{KAIST} & \multicolumn{4}{c}{CVC-14} \\ 
\hline
\multirow{2}{*}{\shortstack{Visible \\ Backbone}} & 
\multirow{2}{*}{\shortstack{Thermal \\ Backbone}} & 
\multirow{2}{*}{\shortstack{Visible \\ Detection Branch}} & 
\multirow{2}{*}{\shortstack{Thermal \\ Detection Branch}} & 
\multicolumn{4}{c|}{$\beta$ = 2} & \multicolumn{4}{c}{$\beta$ = 1} \\ \cline{5-12}

& & & & Algo. 1 & Both & VIS & IR & Algo. 1 & Both & VIS & IR \\ \hline

 & & & & 11.29 & 10.40  & 10.70  & 11.45 & 17.81 & 17.74 & 18.47 & 17.16 \\ 
 $\checkmark$ & $\checkmark$ &   & &  11.40   & 10.52 & 10.56 & 12.25 & 18.64 & 18.61 & 19.06 & 19.75  \\
 
$\checkmark$ & $\checkmark$ &  $\checkmark$ & &   \textbf{8.02} & 8.65 & 8.37 & 10.76 & 17.44 & 18.59 & 17.77 & 21.52 \\
$\checkmark$ & $\checkmark$ &  & $\checkmark$ & 9.58  & 9.48 & 11.51 & 9.85 & \textbf{16.53} & 17.27 & 19.53 & 16.86 \\
 \hline
\end{tabular}
\label{tab:init_and_freeze}
\vspace{-3mm}
\end{table*}
\begin{table}[t!]
\centering
\renewcommand{\arraystretch}{1.2} 
\caption{MR-All results on varying KL Divergence Parameter ($\beta$) for KAIST and CVC-14. Frames and Stride set to 3. } 
\vspace{-0.2cm} 
\resizebox{\columnwidth}{!}{
\begin{tabular}{l|cccc|cccc}
\hline
\multirow{2}{*}{$\beta$} & \multicolumn{4}{c|}{\textbf{KAIST} } & \multicolumn{4}{c}{\textbf{CVC-14} } \\ \cline{2-9}
 &  Algo. 1 & Both & VIS & IR & Algo. 1 & Both & VIS & IR  \\ \hline
0   & 9.84  & 9.66 & 8.51 & 12.29 &  17.26 &   18.23 &  19.38 &      20.23 \\
0.5 &  8.32&     8.95&     8.28&    10.97 &  16.82& 17.77  &18.71  & 17.48  \\
1   &  9.25&    9.96&   8.78&   11.73&       \textbf{16.53} & 17.27& 19.53&  16.86  \\
2   &  \textbf{8.02}&    8.65 &  8.37& 10.76& 17.67  &   18.28&  19.52&  18.61 \\ \hline
\end{tabular}
}
\label{tab:kl_factor}
\vspace{-2mm}
\end{table}

\begin{table}[t!]
\centering
\renewcommand{\arraystretch}{1.2} 
\caption{MR-All with varying $F$ and $S_T$ on KAIST dataset. }
\vspace{-0.2cm} 
\resizebox{\columnwidth}{!}{
\begin{tabular}{c|c|cccc|cccc}
\hline
\multirow{2}{*}{\shortstack{$F$}} & \multirow{2}{*}{$S_T$} & \multicolumn{4}{c|}{\textbf{W/O KL Divergence} ($\beta = 0$)} & \multicolumn{4}{c}{\textbf{With KL Divergence} ($\beta = 2$) } \\ \cline{3-10}
& & Algo. 1 & Both & VIS & IR & Algo. 1 & Both & VIS & IR  \\ \hline
1 & 1  & 9.69   & 9.99  &9.20  &12.63  & 7.53 & 7.44  & 7.93  & 8.82    \\ 
3 & 3  &9.84 & 9.66 & \textbf{8.51} & 12.29 & \textbf{8.02} & 8.65 & 8.37 & 10.76 \\
5 & 3 & 8.84 & 9.00 & 9.09 & 11.04 & 10.01 & 9.70 & 10.04 & 11.39\\ 
7 & 3 & 10.23 & 9.78 & 8.89 & 13.22 & 9.91 & 9.36 & 9.34 & 12.59 \\ 
7 & 10 &  9.34 & 9.29 & 8.79 & 11.95 & 8.39 & 8.49 & 8.79 & 10.24 \\ \hline
\end{tabular}
}
\label{tab:kaist_vary_frame_stride}
\vspace{-2mm}
\end{table}

\begin{table}[t!]
\centering
\renewcommand{\arraystretch}{1.2} 
\caption{ MR-All with varying $F$ and $S_T$ on CVC-14. }
\vspace{-0.2cm} 
\resizebox{\columnwidth}{!}{
\begin{tabular}{c|c|cccc|cccc}
\hline
\multirow{2}{*}{\shortstack{$F$}} & \multirow{2}{*}{$S_T$}  & \multicolumn{4}{c|}{ \textbf{W/O KL Divergence} ($\beta = 0$)} & \multicolumn{4}{c}{\textbf{With KL Divergence} ($\beta = 1$)} \\ \cline{3-10}
                             &    &Algo. 1 & Both & VIS & IR & Algo. 1 & Both & VIS & IR\\ \hline
1 & 1                               & 17.79  & 20.40  & 19.31 & 23.79 & 18.75 & 20.77 & 18.02  & 24.02     \\ 
3 &3                               & 17.26  & 18.23 & 19.38 & 20.23 & 16.53 & 17.27 & 19.53 & 16.86    \\ 
5 &3                              & \textbf{16.76} & 16.96 & 17.78 & 18.48  & \textbf{16.34} & 17.00 & 17.18 & 18.43        \\ 
7 & 3                              & 17.08 & 17.35 & 18.30 & 19.25 & 19.00 & 19.94 & 20.52 & 21.30           \\ 
7 & 5                             & 16.80 & 17.98 & 19.37 & 18.61 & 18.83 & 19.59 & 19.45 & 21.98       \\ \hline
\end{tabular}
}
\label{tab:cvc14_vary_frame_stride}
\vspace{-2mm}
\end{table}

 \begin{table}[h!]
 \centering
 \renewcommand{\arraystretch}{1.2} 
 \caption{MR-All results with and without TAdaConv.}
 \vspace{-0.2cm} 
 \resizebox{\columnwidth}{!}{ 
 \begin{tabular}{c|cccc|cccc}
 \hline
 \multirow{2}{*}{\textbf{Dataset}} & \multicolumn{4}{c|}{W/O TAdaConv} & \multicolumn{4}{c}{With TAdaConv} \\ \cline{2-9}
                                    & Algo. 1 & Both & VIS & IR & Algo. 1 & Both & VIS & IR  \\ \hline
 KAIST                              & 8.59  & 9.81  & 9.23  & 10.40  &  8.02 & 8.65 & 8.37 & 10.76  \\ \hline
 CVC-14                             & 25.85 & 28.47 & 28.2  & 28.17  & 16.47 & 17.15 & 19.62 & 16.80   \\ \hline
 \end{tabular}
 }
 \label{tab:TAdaConv_comparison}
 \vspace{-6mm}
 \end{table}

\begin{table}[t]
    \centering
    \caption{Percentage of Thermal Reliability. \underline{Underline} indicates where there is a contrast with miss rate results.}
        \vspace{-2mm}
    \begin{tabular}{cc}
        \begin{minipage}{0.2\textwidth}
            \subcaption{KAIST}
            \centering
                \vspace{-2mm}
\begin{tabular}{c|c|cc}
                \toprule
                $F$&$S_T$ & $\beta = 0$ & $\beta = 2$ \\
                \midrule
                1 & 1 & \underline{52.70} & \underline{61.38} \\
                3 & 3 & 47.89 & \underline{50.66} \\
                5 & 3 & 48.49 & 49.74 \\
                7 & 3 & 49.21 & \underline{55.79}\\
                7 & 10& 47.70 & \underline{54.21}\\
                \bottomrule
            \end{tabular}
        \end{minipage}
        &
        \begin{minipage}{0.2\textwidth}
            \subcaption{CVC-14}
            \centering
                \vspace{-2mm}
\begin{tabular}{c|c|cc}
                \toprule
                $F$& $S_T$ & $\beta = 0$ & $\beta = 1$\\
                \midrule
                1 & 1 & 39.90 & 31.62 \\
                3 & 3 & 45.43 & \underline{43.21} \\
                5 & 3 & 47.13 & 46.12 \\
                7 & 3 & 44.16 & 45.60\\
                7 & 5 & \underline{45.25} & 44.95\\
                \bottomrule
            \end{tabular}
        \end{minipage}
    \end{tabular}    
\label{tab:reliablity_cvc_kaist}
\vspace{-6.5mm}
\end{table}

\subsection{Results Compared with State-of-the-Art}
We compare our approach against eight state-of-the-art multimodal methods evaluated on both datasets: AR-CNN \cite{zhang2019weakly}, CMM \cite{kim2024causal}, MBNet\cite{zhou2020improving}, Kim et al. \cite{kim2021uncertainty}, MLPD\cite{kim2021mlpd}, MambaST \cite{gao2024mambast}, Lee et al. \cite{lee2022cross}, and MS-DETR \cite{xing2024ms}. Among these, AR-CNN, MS-DETR, MBNet, Kim et al. \cite{kim2021uncertainty}, and CMM are specifically designed to address misalignment and modality imbalance. MambaST and Lee et al. \cite{lee2022cross} use Mamba and Transformer-based fusion strategies. MLPD focuses on handling overlapping and partially overlapping image pairs.

\begin{figure*}[h]
\centering
\includegraphics[width=.9\textwidth]{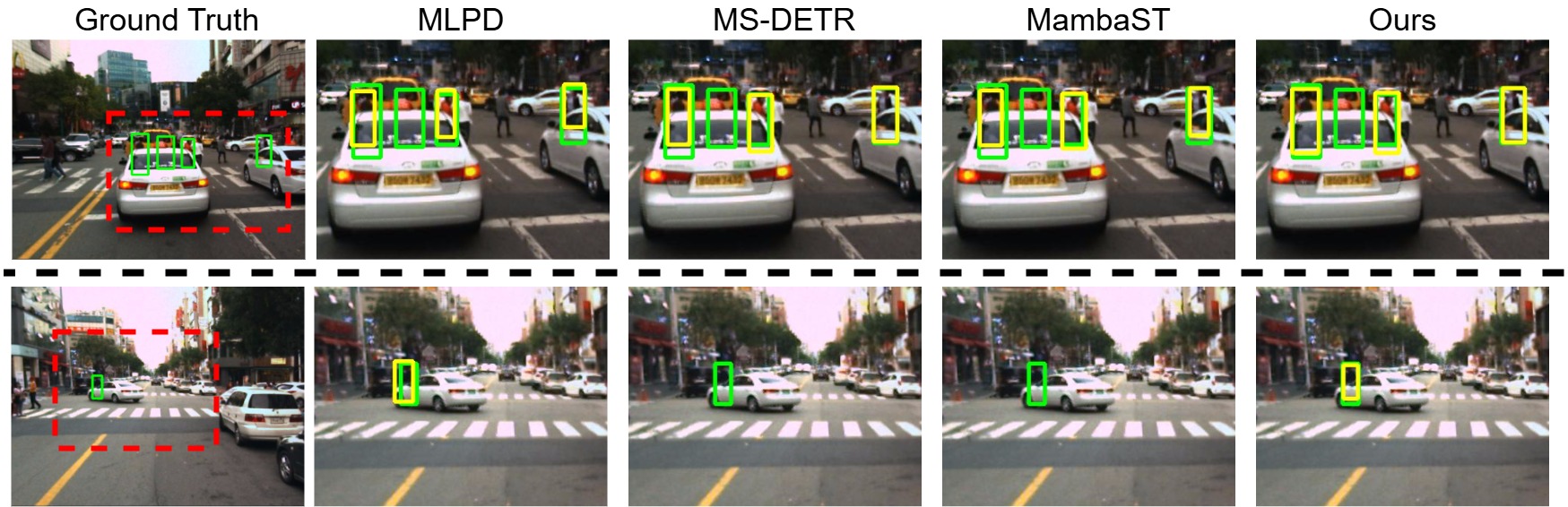}
\caption{\textcolor{black}{Visual examples of KAIST detection results for heavily occluded pedestrians. Green and yellow bounding boxes correspond to ground truth and detection results, respectively. Ours show better detection on occluded/small pedestrians. Note that the KAIST dataset contains well aligned examples so we show only the visible images \textcolor{black}{(See Multimedia
Fig. R1).}}} 
\label{fig:KAIST_image}
\vspace{-3mm}
\end{figure*}

\textcolor{black}{On the KAIST dataset, Table~\ref{table:mr_KAIST_cvc_14} shows our results are on par with leading methods (MambaST and MS-DETR), while outperforming AR-CNN, MBNet, and MLPD. Compared with MS-DETR, our spatial-temporal method, Strip-Fusion, is more effective for heavily occluded pedestrians. Specifically, Table~\ref{table:additional_kaist} shows that Strip-Fusion achieves the best ``MR-All" of 41.05 under the ``Heavy Occlusion” setting, with qualitative comparisons illustrated in Fig.~\ref{fig:KAIST_image}. MS-DETR and MambaST slightly outperform our method on medium-sized and partially occluded pedestrians. For ``far" pedestrians, our method achieves the second-best performance on KAIST, likely because avoiding down-sampling in the fusion block preserves fine-grained features for small-scale pedestrians. }

\textcolor{black}{On the CVC-14 dataset, Table~\ref{table:mr_KAIST_cvc_14} shows we achieve state-of-the-art ``MR-All" and ``MR-Day" scores of 16.34 and 19.30, respectively. Additionally, as shown in Table~ \ref{tab:additional_cvc14}, Strip-Fusion outperforms MS-DETR on larger-scale pedestrians, indicating that our concatenation and addition-based fusion combined with post-processing remains robust to misalignment for ``larger" pedestrians. In contrast, on CVC-14, MS-DETR achieves superior performance on far pedestrians ($<45$ pixels), likely due to its Deformable Transformer-based fusion, which appears to better align small-scale pedestrian features.
}

As for inference time, in our experiments, Strip-Fusion performed between ~2.7 fps given shorter (3-frame) sequences and ~1.6 fps for longer (7-frame) sequences \textcolor{black}{(See Multimedia Tables R3 \& R4)}.

\subsection{Ablation Studies} 
\label{sec:ablation}
To better understand the contributions of individual components in our method, we conducted a series of ablation studies, including investigating the impact of freezing of base weights, varying the number of frames and stride, incorporating TAdaConv, KL divergence, and post-processing strategy. 

\begin{figure*}[h]
\centering
\includegraphics[width=.9\textwidth]{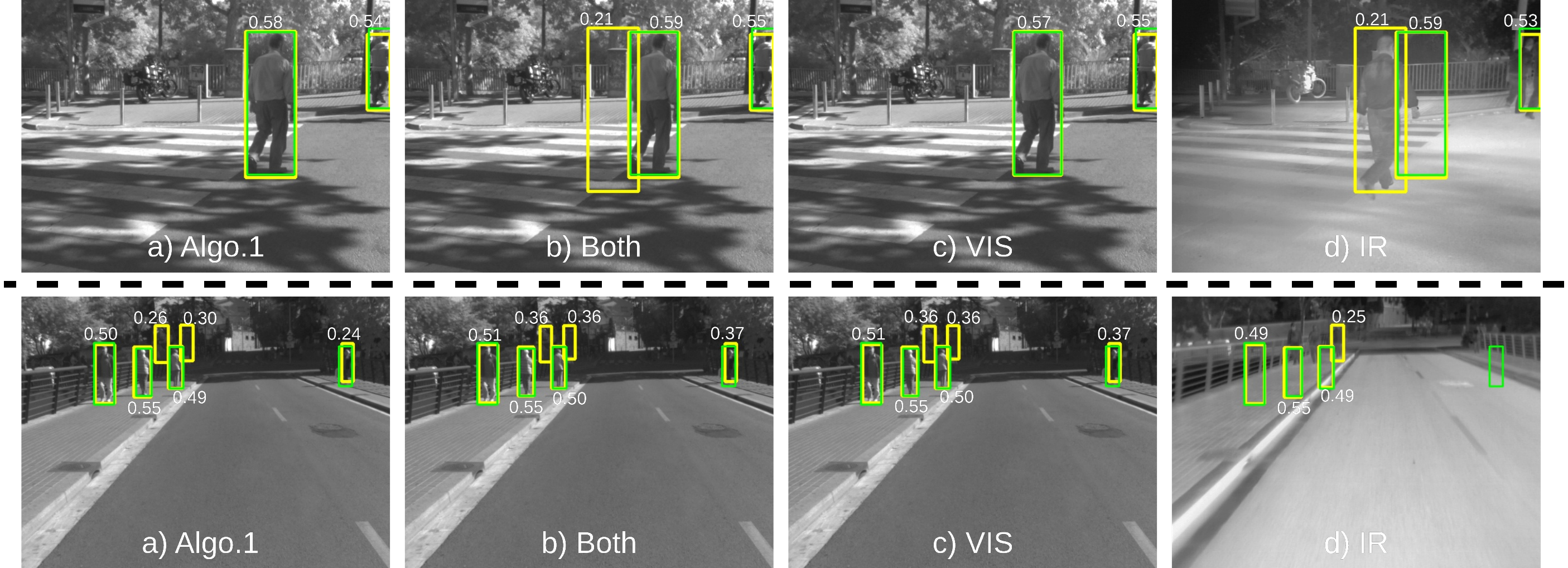}
\caption{ 
Visual results on the CVC-14 dataset for two misaligned image pairs. Numbers indicate detection confidence scores. The top row shows that our Algo.1 post-processing approach reduces false positives by removing low-confidence and non-overlapping detections, leading to more consistent performance. The bottom row illustrates a ``failure'' case where high-confidence false positives persist after post-processing in areas of misalignment. Detections below 0.2 confidence are not shown.
}
\label{fig:cvc14_images}
\vspace{-5mm}
\end{figure*}

In the following, we experimented with four post-processing techniques. Sequences of visible and thermal images were first processed by the visible and thermal backbone, and passed through our proposed strip fusion module. The first approach applies Non-Maximum Suppression (NMS) to the visible detection branch and its corresponding head, noted as ``VIS". The second applies NMS solely to the thermal detection branch and head, noted as ``IR". The third performs joint NMS across both visible and thermal branches and their outputs, noted as ``Both". The fourth approach also uses both branches, but differs by first applying our newly proposed post-processing strategy (our Algorithm~\ref{alg:fuse_predictions}) before joint NMS, noted as ``Algo. 1". 
We provide numerical ablation results across all four strategies on the ``reasonable" setting, as well as visual illustration on the impact of post processing in Fig.~\ref{fig:cvc14_images}.

\subsubsection{Freezing base weights}
\label{sec:freezebaseweights}
 Table~\ref{tab:init_and_freeze} supports our optimization choices and justifies freezing the base weights. Temporally adaptive convolutions (TAdaConv) \cite{huang2021tada} were used in the modality-specific backbones and detection branches in our pipeline. The original TAdaConv work \cite{huang2021tada} discussed facilitating learning spatial-temporal features by optimizing both the calibration and base weights. In our models, we find that freezing the base weights of both backbones (visible and thermal) and one detection head (visible or thermal), while learning their calibration weights, leads to better results (see Table~\ref{tab:init_and_freeze}). However, freezing base weights tends to increase the performance gap between the VIS and IR columns, making a post-processing strategy such as  Algorithm~\ref{alg:fuse_predictions} necessary. When base weights are unfrozen, the outputs from both detection heads become more similar in terms of ``MR-All", but the overall performance drops.

\subsubsection{KL Divergence}
Table \ref{tab:kl_factor} reinforces our optimization settings for the KL divergence hyperparameter $\beta$ across both datasets in our default training setup, where $F$ and $S_T$ are set to 3. For the four post-processing settings in Table \ref{tab:kl_factor}, we observe that, on both datasets, a majority of the miss rate values in each column improve when using KL divergence compared to when $\beta = 0$ (without KL divergence). This further demonstrates that incorporating KL divergence enhanced the performance of our proposed method, particularly for shorter temporal windows.

\subsubsection{Varying Number of Frames ($F$) and Stride ($S_T$)} 
Tables \ref{tab:kaist_vary_frame_stride} and \ref{tab:cvc14_vary_frame_stride} present the results for the KAIST and CVC-14 datasets as the sequence length increases.
Both tables indicate that increasing the temporal window does not monotonically improve the miss rate. For models using KL divergence, this is likely because the ROI features for a sequence were extracted using the current frame's bounding box. This led to competitive results for shorter temporal windows but does not consistently improve performance for longer temporal windows likely due to increased probability of significant movement by both the pedestrians and the ego vehicle. From Table \ref{tab:cvc14_vary_frame_stride}, we observe that models without KL divergence, achieved better performance for longer time windows ($F/S_t = 7/3$ and $F/S_t = 7/5$) across the four post-processing approaches compared to using KL divergence.

\subsubsection{TAdaConv}
The original TAdaConv assumes that the input channel and output channels are equal. We relaxed this constraint and modified the dimensional reduction for the 1-D convolutions and the fully connected layer. (See section \ref{sec:freezebaseweights} for a brief explanation of how we use TAdaConv and freeze base weights). Table~\ref{tab:TAdaConv_comparison} shows the ablation results with and without TAdaConv in our fusion pipeline, and we observe that incorporating TAdaConv improved performance on both datasets, with a significant gain observed on CVC-14. This improvement can be attributed to TAdaConv’s calibration weights, which allow the model to more effectively capture and integrate spatial-temporal information. The improvement is also influenced by our training setup, which involves freezing both backbones and one modality during training.

\subsubsection{Thermal Reliability Percentage}
The reliability percentage is an indicator of how closely a modality's average results match the ground truth. This provides an explainable metric to justify the effectiveness of our KL divergence loss. For each image and scale, we compute the reliability for both modalities—\textcolor{black}{visible ($r^v$) and thermal ($r^t$)}. If $r^t > r^v$, the thermal count is increased by one. If neither modality has overlapping objects with the ground truth at a particular scale, that instance is excluded from the count. The thermal reliability percentage is then calculated by dividing the thermal count by the total number of valid instances. The visible reliability percentage is simply the complement (1 minus) of the thermal reliability percentage.

We report the thermal reliability percentage during inference on the KAIST and CVC-14 datasets in  Table~\ref{tab:reliablity_cvc_kaist}. Interestingly, we observed a contrast between reliability percentage and miss rate results in Tables~\ref{tab:kaist_vary_frame_stride} and~\ref{tab:cvc14_vary_frame_stride}, particularly in the VIS and IR columns. \textcolor{black}{Specifically, in Table \ref{tab:kaist_vary_frame_stride} with $F, S_T = 1, 1$, the VIS miss rate is 7.93 (lower) and the IR miss rate is 8.82 (higher), yet the thermal reliability is 61.38\%, indicating the algorithm thinks thermal is slightly more ``reliable''. We believe the contrast was caused by a few well-aligned, high-confidence visible detections that reduced the miss rate.}
\section{Conclusion}
\vspace{-1mm}
This paper presents \textit{Strip-Fusion}, a spatial-temporal fusion model for multispectral pedestrian detection. \textcolor{black}{\textit{Strip-Fusion} fuses visible and thermal image pair sequences through several innovations, including incorporating TAdaConv to implicitly learn temporal features, a novel strip fusion module for spatial-temporal fusion, a novel KL divergence loss addressing modality imbalance, and a novel post processing strategy.} Even though our method does not perform explicit alignment, we show competitive performance on both KAIST and the CVC-14 benchmarks with spatial misalignment. 
  
\small
\bibliographystyle{IEEEtran}
\bibliography{references}

\end{document}